\documentclass[conference]{IEEEtran}
\IEEEoverridecommandlockouts
\usepackage{cite}
\usepackage{amsmath,amssymb,amsfonts}
\usepackage{algorithmic}
\usepackage{graphicx}
\usepackage{textcomp}
\usepackage{xcolor}

\usepackage{cite}
\usepackage{soul,color}
\usepackage{amsmath,amssymb,amsfonts}
\usepackage{graphicx}
\usepackage{subcaption}
\usepackage{textcomp}
\usepackage{xcolor}
\usepackage{booktabs}
\usepackage{comment}
\usepackage{hyperref}
\usepackage{multirow}

\def\BibTeX{{\rm B\kern-.05em{\sc i\kern-.025em b}\kern-.08em
    T\kern-.1667em\lower.7ex\hbox{E}\kern-.125emX}}
\begin{document}

\title {Multimodal Gender Fairness in Depression Prediction: Insights on Data from the USA \& China}

%\author{Anonymous}

\author{\IEEEauthorblockN{Joseph Cameron}
\IEEEauthorblockA{
\textit{University of Cambridge}\\
Cambridge, UK \\
jmc276@cam.ac.uk}
\and
\IEEEauthorblockN{Jiaee Cheong}
\IEEEauthorblockA{
\textit{University of Cambridge,}\\
\textit{the Alan Turing Institute}\\
jc2208@cam.ac.uk}
\and
\IEEEauthorblockN{Micol Spitale}
\IEEEauthorblockA{
\textit{University of Cambridge}\\
Cambridge, UK \\
ms2871@cam.ac.uk}
\and
\IEEEauthorblockN{Hatice Gunes}
\IEEEauthorblockA{
\textit{University of Cambridge}\\
Cambridge, UK \\
hg410@cam.ac.uk}
}

\begin{comment}

\author{\IEEEauthorblockN{1\textsuperscript{st} Given Name Surname}
\IEEEauthorblockA{\textit{dept. name of organization (of Aff.)} \\
\textit{name of organization (of Aff.)}\\
City, Country \\
email address or ORCID}
\and
\IEEEauthorblockN{2\textsuperscript{nd} Given Name Surname}
\IEEEauthorblockA{\textit{dept. name of organization (of Aff.)} \\
\textit{name of organization (of Aff.)}\\
City, Country \\
email address or ORCID}
\and
\IEEEauthorblockN{3\textsuperscript{rd} Given Name Surname}
\IEEEauthorblockA{\textit{dept. name of organization (of Aff.)} \\
\textit{name of organization (of Aff.)}\\
City, Country \\
email address or ORCID}
\and
\IEEEauthorblockN{4\textsuperscript{th} Given Name Surname}
\IEEEauthorblockA{\textit{dept. name of organization (of Aff.)} \\
\textit{name of organization (of Aff.)}\\
City, Country \\
email address or ORCID}
\and
\IEEEauthorblockN{5\textsuperscript{th} Given Name Surname}
\IEEEauthorblockA{\textit{dept. name of organization (of Aff.)} \\
\textit{name of organization (of Aff.)}\\
City, Country \\
email address or ORCID}
\and
\IEEEauthorblockN{6\textsuperscript{th} Given Name Surname}
\IEEEauthorblockA{\textit{dept. name of organization (of Aff.)} \\
\textit{name of organization (of Aff.)}\\
City, Country \\
email address or ORCID}
}
\end{comment}

\maketitle

\begin{abstract}
\begin{comment}

The problem of bias and fairness in machine learning (ML) algorithms is becoming an increasingly greater source of concern. Existing work has indicated that ML bias is present across gender for the task of depression detection. However, none of the existing work has taken into consideration the difference in depression manifestation across different countries. We hypothesise that the representation of features (acoustic, textual, and visual) and their inter-modal relations would significantly vary among subjects from different countries, thus impacting the performance and fairness of various ML models. We present the very first evaluation of multimodal gender fairness in depression manifestation by undertaking a study on data from the USA and China. We undertake a thorough statistical and ML experimentation, and evaluate the performance and fairness of various machine learning models trained on the aforementioned datasets. 
%
Our findings indicate that there may indeed be a difference in depression manifestation across countries and cultures (and other underlying factors) which may have exacerbated the identified differences. Our findings further motivate a call for more consistent data collection process in order to address the problem of ML bias in depression detection.
\end{comment}

Social agents and robots are increasingly being used in wellbeing settings. 
However, a key challenge is that these agents and robots typically rely on machine learning (ML) algorithms to detect and analyse an individual's mental wellbeing.
The problem of bias and fairness in ML algorithms is becoming an increasingly greater source of concern. 
%
%Existing work has indicated that ML bias is present across gender for the task of depression detection.
%
In concurrence, existing literature has also indicated that mental health conditions can manifest differently across genders and cultures.
%
%None of the existing work has taken into consideration the difference in depression manifestation across different countries. 
%
We hypothesise that the representation of features (acoustic, textual, and visual) and their inter-modal relations would vary among subjects from different cultures and genders, thus impacting the performance and fairness of various ML models. 
We present the very first evaluation of multimodal gender fairness in depression manifestation by undertaking a study on two different datasets from the USA and China. We undertake thorough statistical and ML experimentation and 
%evaluate the performance and fairness of various ML models trained on the aforementioned datasets.
%
%made a conscious effort to 
repeat the
experiments for several different algorithms to ensure that the results
are not algorithm-dependent.
%
%Our findings indicate that there is a statistically significant difference across some features 
Our findings indicate that though %differences exist across both performance and fairness measures, 
there are differences between both datasets, it is not conclusive whether this is due to the difference in depression manifestation as hypothesised or other external factors such as differences in data collection methodology.
%and identified challenges that hampered conclusive findings across both datasets.
%
%
%may indeed be a difference in depression manifestation across countries and cultures (and other underlying factors) which may have exacerbated the identified differences. 
%
Our findings further motivate a call for a more consistent and culturally aware data collection process in order to address the problem of ML bias in depression detection and to promote the development of fairer agents and robots for wellbeing.

\end{abstract}


%%%%%%%%%%%%%%%%%%%%%%%%%%%%%%%%%%%%%%%%%%%%%%%%%%%%%%%%%%%%%%%%%%%%%%%%%%%%
%%%%%%%%%%%%%%%%%%%%%%%%%%%%%%%%%%%%%%%%%%%%%%%%%%%%%%%%%%%%%%%%%%%%%%%%%%%%

%%%%%%%%%%%%%%%%%%%%%%%%%%%%%%%%%%%%%%%%%%%%
%
\section{Introduction}

Mental health disorders (MHDs) are becoming increasingly prevalent \cite{maj2020clinical}.
In concurrence, there is a growing body of research which started to explore the use of agents or robotic coaches to support mental wellbeing \cite{jeong2023deploying, spitale2023vita, axelsson2023robotic,spitale2024appropriateness}. 
%Machine learning (ML) methods have been applied to many health-related areas \cite{sendak2020human}. 
%
%The field of machine learning offers powerful tools for analyzing and predicting mental health states within humans from both unimodal and multimodal data sources  \cite{su2020deep,shatte2019machine}. 
%
%
%
%
%%%%%%%%%%%%%%%%%%%%%%%%%%%%%%%%%%%%%%%%%%%%
However, a unique challenge is that mental health conditions
%, particularly depression, 
manifest 
%and show symptoms 
in diverse ways across individuals. The different manifestations and symptoms can be influenced by gender \cite{barsky2001somatic,ogrodniczuk2011men,floyd1997problems}
and countries \cite{Juhasz_cultural_2012}.
%
%Across country, 
Brody et al. show that women aged 20 or over are nearly twice as likely to show signs of depression when compared to men aged 20 or over \cite{Brody_prevalence_2018} and Hasin et al. found a similar pattern in their survey of American adults (aged over 18) that targeted an investigation of major depressive disorders \cite{hasin_epidemiology_2018}. 
Chang et al. observed that Koreans were far more likely to show symptoms of low energy or difficulties with concentration, whereas Americans were less likely to show concentration difficulty and far more likely to show high levels of work productivity \cite{Chang_crossnational_2008}. 
Following such insights, Juhasz et al. identify that depression can manifest and be displayed in very different ways across different nations and cultures \cite{Juhasz_cultural_2012}.
As a result, many of the western-developed depression diagnostic tools 
%and frameworks 
may be biased towards and better at identifying depression within Northern American and European demographic groups \cite{Juhasz_cultural_2012}.
%reflections of what depression is \cite{Juhasz_cultural_2012}.

%%%%%%%%%%%%%%%%%%%%%%%%%%%%%%%%%%%%%%%%%%%%
Concurrently, ML bias is becoming a growing source of concern \cite{barocas2017fairness,cheong2021hitchhiker,spitale2024underneath,cheong2023counterfactual,xu2020investigating}.
Given the high stakes involved in MHD analysis, it is crucial to investigate and mitigate the ML biases present. 
Recent research has indicated that ML bias is present in existing MH datasets and models \cite{bailey2021gender,zanna2022bias,cheong_gender_fairness}.
However, none of the existing work has investigated the problem of ML bias across different countries and cultures. 
The gender and cultural differences in depression manifestation, diagnosis, and general symptoms described in \cite{Chang_crossnational_2008,Juhasz_cultural_2012,Brody_prevalence_2018}
clearly communicate that symptoms of depression can significantly differ between different genders and nations or cultures, potentially leading to biased frameworks and datasets for depression diagnosis and manifestation.
%
%the fairness of these machine learning models across different demographic groups remains a critical concern, as biases in training data or algorithms can lead to biased care or involuntary misdiagnosis\cite{bailey2021gender, park2022fairness,park2021comparison,cheong_gender_fairness}.
%
%%%%%%%%%%%%%%%%%%%%%%%%%%%%%%%%%%%%%%%%%%%%
%
Thus, we hope to tackle the aforementioned gaps by addressing the following research questions (RQs): 

\textbf{RQ 1:} Are there any differences in depression manifestation across gender and countries? If present, what are the primary sources of difference in depression manifestation? 
To address this research question, we conduct thorough statistical analysis of features between two datasets collected from different countries, namely China and the USA, and evaluate to what extent the difference in features impacts model performance.
\textbf{RQ 2:} How do the results compare between different ML models and modalities? 
To address this research question, we conduct thorough experimentation using a range of ML models to evaluate how the different models perform across the different datasets and modalities. 
\textbf{RQ 3:} Are there any differences in ML model performance and fairness across gender and countries? If present, are the differences in performance due to the difference in depression manifestation as hypothesised? 
To address this research question, we draw on the findings from the previous two RQs and further evaluate some key insights that may have impacted performance and fairness results.

The main contributions of this paper are as follows. First, we aim to understand how biases occur with respect to gender and whether they differ based on the location and dataset curation method.
We do so by conducting experiments on two multimodal
%international 
depression  %corpora 
datasets from the USA and China.
Next, we evaluate the performance and fairness of various machine learning models trained on the aforementioned datasets.
%
%The goal of our research 
%Our goal is to understand if the algorithms are truly biased or simply ill-suited to adapt towards the different manifestation of depression across various countries and cultures.
%, helping to ensure equitable mental health diagnosis and assessments for all individuals in the future, irrespective of their gender or nationality.
%

%%%%%%%%%%%%%%%%%%%%%%
% LITERATURE REVIEW

\section{Literature Review}

%%%%%%%%%%%%%%%%%%%%%%%%%%%%%%%%%%%%

\subsection{Machine Learning in Depression Prediction}
There have been recent attempts at using machine learning methods, including both unimodal and multimodal approaches to automatically analyse and predict depression using extracted features \cite{poria2017review,benssassi2021investigating,cheong2024fairrefuse}. 
It is possible to do so via several different data sources such as physiological data \cite{yau2022tiles,mundnich2020tiles}, motor activity data \cite{jakobsen2020psykose,garcia2018depresjon} or audio-visual sources \cite{he2022deep,Yoon_Kang_Kim_Han_2022,zhang2020multimodal}. Audio-visual (AV) datasets typically include behavioural signals such as facial affect, body gestures and vocal intensity \cite{devault2014simsensei,gratch2014distress,he2022deep}. 
In general, multimodal approaches are shown to perform better than unimodal approaches \cite{cohn2018multimodal,uddin2022deep,wei2022multi}. 
Most recently, Zou \textit{et al.} explored the effects of using unimodal and multimodal variations of the visual, acoustic, and textual features extracted from the CMDC dataset \cite{zou_cmdc_2023}.
The visual features Zou \textit{et al.} identified for extraction from the CMDC dataset were Eye Gazes, Eye Landmarks, Head Poses, Facial Landmarks and Facial Action Units (AU) \cite{ekman_1978_facial}.
%
%These visual features can all be major communicators of major depressive disorder \cite{Ellgring1989-mw}.
%
%Pampouchidou \textit{et al.} also showed much promise when applying facial action units \cite{ekman_1978_facial} to automatically predict depression from visual cues \cite{pampouchidou_automatic_2019}.
%

%Acoustic features that were extracted by Zou \textit{et al.} include Mel Frequency Cepstral Coefficients (MFCC) and follow GeMAPS laid out by Eyben \textit{et al.} \cite{eyben_eGeMAPS_2016}.
%
%Zou \textit{et al.} perform statistical analysis via Multivariate Analysis of Covariance (MANCOVA) to compare differences between the visual, acoustic, and textual feature modalities for both non-depressed participants and participants with major depressive disorder \cite{zou_cmdc_2023}.
%
%Then, Zou \textit{et al.} perform benchmark evaluations with different machine learning algorithms being trained on all the different combinations of visual, acoustic, and textual features in unimodal and multimodal terms to compare the performance metrics of each machine learning algorithm \cite{zou_cmdc_2023}.

%%%%%%%%%%%%%%%%%%%%%%%%%%%%%%%%%%%%

\subsection{Difference in Manifestation of Depression}
%\textcolor{red}{New}
%
Existing literature indicates that depression can manifest in vastly different ways across different cultures, countries and genders \cite{Chang_crossnational_2008,ogrodniczuk2011men,barsky2001somatic}. Brody \textit{et al.} and Hasin \textit{et al.} demonstrate that adult women are more likely to display signs of depression \cite{Brody_prevalence_2018,hasin_epidemiology_2018}. Platt \textit{et al.} investigate this observation further and attribute the increased manifestation of depression in women to the gender pay gap \cite{Platt_unequal_2016}. 
%Salk \textit{et al.} also highlight how nations with higher gender equity also had a higher gender difference in depression diagnoses \cite{Salk_gender_2017}.
%
Across cultural
%international 
differences in depression symptoms, Chang \textit{et al.} observed a difference in communicated energy levels and concentration between depressed Koreans and Americans
%, where depressed Koreans were more likely to display symptoms of low energy and concentration problems compared to Americans 
\cite{Chang_crossnational_2008}. 
Further, Iwata and Buka found that 
%when levels of positive affect between Japanese university students and American university students were measured, 
Japanese university students reported higher levels of low positive affect compared to American university students \cite{Iwata_race_2002}. 
As a result, Japanese students were considered to be more depressed 
%on the western–developed Center for Epidemiological Studies Depression Scale (CES-D) \cite{Radloff_cesd_1977} 
than their American counterparts although this may not necessarily be true \cite{Iwata_race_2002}.

%%%%%%%%%%%%%%%%%%%%%%%%%%%%%%%%%%%%

\subsection{ML Fairness in Mental Wellbeing }
%\textcolor{red}{New:}
%
There is only a handful of studies which have looked into bias in mental well-being prediction \cite{ryanfairness, bailey2021gender, park2022fairness, park2021comparison, zanna2022bias,cheong_gender_fairness,Cheong_NotFair_2023}.
Park \textit{et al.} \cite{park2021comparison} conducted their experiments on data collected in a clinical setting with a specific focus on post-partum depression. 
Zanna \textit{et al.} \cite{zanna2022bias} conducted their experiments on data collected in the wild with a specific focus on anxiety prediction. 
%
%Cheong \textit{et al.} \cite{cheong_gender_fairness} examined whether bias exists in existing mental health datasets and algorithms and provided practical suggestions to avoid hampering bias mitigation efforts in ML for mental health.
%
%Ryan \textit{et al.} \cite{ryanfairness} proposed three categories of fairness definitions they deem relevant to mental health. 
%
Park \textit{et al.} \cite{park2022fairness} analysed bias across gender in mobile mental health assessment and proposed an algorithmic impact remover to mitigate unwanted bias.
Bailey and Plumbley \cite{bailey2021gender} attempted to mitigate the gender bias present in the DAIC-WOZ dataset using data re-distribution.
Cheong \textit{et al.} evaluated the impact of gender bias on depression prediction 
%within the British AFAR-BSFT dataset 
with various ML models using various unimodal and multimodal approaches \cite{Cheong_NotFair_2023}.

However, all the existing works have chiefly focused on investigating ML fairness for multiple ML models across gender using a dataset originating from \textit{one} country.
There are no known comparative evaluations of ML fairness for mental health using unimodal and multimodal approaches across different countries or cultures.
More specifically, none of the existing works have investigated the problem of ML bias across different countries and/or cultures. 
The novel contribution of this paper is to present a thorough comparative evaluation of the performance and fairness of various ML models using unimodal and multimodal approaches when trained on the American E-DAIC dataset \cite{gratch-etal-2014-distress} and the Chinese CMDC dataset \cite{zou_cmdc_2023}, thus allowing for the comparison of ML fairness across gender and countries and across multiple modalities.

\section{Experimental Setup}

In this section, we provide the dataset information, and implemented data pre-processing and model training details.

%%%%%%%%%%%%%%%%%%%%%%%%%%%%%%%%%%%%%%%%%%%%%%%%

\subsection{Datasets \& Mental Health Measurements}

We use the Chinese Multimodal Depression Corpus (CMDC) dataset \cite{zou_cmdc_2023} and
the American Extended Distress Analysis Interview Corpus (E-DAIC) dataset \cite{gratch-etal-2014-distress} to address our research questions. 
%They both contain textual, visual, and acoustic data
%
%%%%%%%%%%%%%%%%%%%%%%%%%%%%%%%%%%%%%%%%%%%%%%%%
%\subsubsection{The CMDC Dataset}
%
The \textbf{CMDC} \cite{zou_cmdc_2023} is a Chinese dataset containing textual, acoustic, and visual data collected from
%It contains 
%audio and video recordings of semi-structured interviews 
%between the authors and 
78 participants. %, where 
The semi-structured interview scripts contained questions and topics of discussion verified by clinicians.
%
%CMDC contains textual, acoustic, and visual data. 
%
%78 participants contributed acoustic and textual features for further analysis and 
%45 participants contributed additional visual features for further analysis. 
%
%Table \ref{tab:CMDCOverviewOfParticipants} shows an overview of the participants and their attributes. %across the CMDC dataset.
%
%CMDC relies on the 
Ground-truth labels were obtained using the
Patient Health Questionnaire-9 (PHQ-9)
%to diagnose for depression 
\cite{Kroenke_PHQ8_2009}.
%
%
%
\begin{comment}

\begin{table}
\begin{center}
  \caption{Overview of Participants in the CMDC Dataset}
  \label{tab:CMDCOverviewOfParticipants}
  \begin{tabular}{cccccc}
    \toprule
    \textbf{Recording} & \multicolumn{2}{c}{\textbf{MDD}} & \multicolumn{2}{c}{\textbf{HC}} & \textbf{Total}\\
     & Male & Female & Male & Female & \\
    \midrule
    Audio & 8 & 18 & 17 & 35 & 78\\
    Video + Audio & 5 & 14 & 6 & 20 & 45\\
  \bottomrule
\end{tabular}
\end{center}
\end{table}
\end{comment}
%
%%%%%%%%%%%%%%%%%%%%%%%%%%%%%%%%%%%%%%%%%%%%%%%%
%\subsubsection{The E-DAIC Dataset}
The \textbf{E-DAIC} \cite{ringeval_AVEC_2019, devault_DAICvirtualinterviewer_2014} is a US-based dataset. 
%Similar to CMDC, 
E-DAIC contains textual, acoustic and visual data 
%This was 
collected from 275 participants.
%
%Table \ref{tab:EDAICOverviewOfParticipants} shows an overview of the participants and their attributes across the E-DAIC dataset.
%
%sA key difference is that 
E-DAIC employs the Patient Health Questionnaire-8 (PHQ-8) to obtain the ground-truth label \cite{Kroenke_PHQ8_2009}.
%as a measure of how depressed participants were via a total PHQ score \cite{Kroenke_PHQ8_2009}.
%
The only difference between the PHQ-8 and the PHQ-9 is that the PHQ-8 does not include the ninth item which relates to suicidal thoughts. 

%%%%%%%%%%%%%%%%%%%%%%%%%%%%%%%%%%%%%%%%%%%%%%%%
\subsection{Data Pre-Processing}
% -----------
\subsubsection{CMDC Dataset}
\label{subsubsec:CMDCMissingData}
%Table \ref{tab:CMDC_Missing_Visual_Data} shows the missing visual data , Table \ref{tab:CMDC_Missing_Acoustic_Data} shows the missing acoustic data, and Table \ref{tab:CMDC_Missing_Textual_Data} shows the missing textual data in the CMDC dataset.
%
%
%As a result of the above missing data, 
The dataset provided ready to use visual features such as action units (AUs), eye gaze, head pose and facial landmarks extracted using OpenFace (version 2.2.0) \cite{Baltrusaitis_OpenFace_2018}.
OpenFace is used because it is considered the state-of-the-art open-sourced tool for facial analysis. 
AUs are used as extracted features to describe the changing characteristics of facial expression in depression since depressed
patients may exhibit poor expression ability. 
Moreover, the
occurrence of specific facial movements (e.g. smile, corner of
mouth down, etc.) described by specific AUs (e.g., AU12) is
directly related to depression.
The acoustic features set used is the extended Geneva
Minimalistic Acoustic Parameter Set (eGeMAPS) \cite{eyben_eGeMAPS_2016}.
Another widely used open-source toolkit, openSMILE (version 3.0) \cite{eyben_OPENSMILE_2010}, was used for acoustic feature extraction. Some examples of extracted acoustic features include pitch, jitters,
frequency, and bandwidth of Formant 1, 2, 3 etc.
Visual data for questions 2, 8, 10, 11, and 12 for all participants were removed 
%when performing further analysis and 
from subsequent ML experimentation 
as they were missing.
%
%This was necessary to maintain a representative fixed-length visual feature vector for subsequent ML experimentation between all participants. 
%
%It was unfeasible to replace missing values with traditional imputation techniques due to the variation of missing values across participants.
%
\begin{comment}
%
\begin{table}
  \caption{Missing Visual Data in the CMDC Dataset}
  \label{tab:CMDC_Missing_Visual_Data}
  \begin{tabular}{ccc}
    \toprule
    \textbf{Participant} & \textbf{Gender} & \textbf{Missing Question Data}\\
    \midrule
    HC 24 & Female & Q10\\
    HC 25 & Female & Q2\\
    HC 49 & Female & Q12\\
    MDD 02 & Female & Q11, Q12\\
    MDD 20 & Female & Q8\\
    MDD 21 & Female & Q12\\
  \bottomrule
\end{tabular}
\end{table}
%
\end{comment}
%
%We removed acoustic data for interview questions 5 and 8 and textual data for interview questions 6, 8, 9, 10, 11, and 12 for all participants for the same reason mentioned previously.
%
In addition, we remove participant MDD 23 %from the textual data modality 
as they had no representative textual data.
%
\begin{comment}
%
\begin{table}
  \caption{Missing Acoustic Data in the CMDC Dataset}
  \label{tab:CMDC_Missing_Acoustic_Data}
  \begin{tabular}{ccc}
    \toprule
    \textbf{Participant} & \textbf{Gender} & \textbf{Missing Question Data}\\
    \midrule
    MDD 05 & Female & Q8\\
    MDD 07 & Female & Q5\\
  \bottomrule
\end{tabular}
\end{table}
%
%
\begin{table}
  \caption{Missing Textual Data in the CMDC Dataset}
  \label{tab:CMDC_Missing_Textual_Data}
  \begin{tabular}{ccc}
    \toprule
    \textbf{Participant} & \textbf{Gender} & \textbf{Missing Question Data}\\
    \midrule
    HC 24 & Female & Q10\\
    HC 40 & Female & Q10\\
    HC 41 & Female & Q9, Q10, Q11, Q12\\
    HC 49 & Female & Q12\\
    MDD 01 & Female & Q12\\
    MDD 11 & Female & Q6, Q9, Q10\\
    MDD 14 & Female & Q11, Q12\\
    MDD 17 & Male & Q9, Q11\\
    MDD 20 & Female & Q8\\
    MDD 23 & Male & All Qs\\
  \bottomrule
\end{tabular}
\end{table}
%
\end{comment}
%
Zou et al. did not mention any of the identified gaps in the CMDC dataset \cite{zou_cmdc_2023}. 
%It is possible that some participants may have subsequently requested certain aspects of their data to be deleted following the initial release of CMDC.

% -----------
\subsubsection{E-DAIC Dataset}
\label{sect:edaic_preprocessing}

\begin{comment}

\begin{table}
  \caption{Overview of Participants in the E-DAIC Dataset}
  \label{tab:EDAICOverviewOfParticipants}
  \begin{tabular}{cccccc}
    \toprule
    \textbf{Recording} & \multicolumn{2}{c}{\textbf{MDD}} & \multicolumn{2}{c}{\textbf{HC}} & \textbf{Total}\\
     & Male & Female & Male & Female & \\
    \midrule
    Video + Audio & 35 & 31 & 135 & 74 & 275\\
  \bottomrule
\end{tabular}
\end{table}
\end{comment}

There was no missing data in E-DAIC.
For the fairest comparison between CMDC and E-DAIC, we pre-process the data in a similar fashion as done by Zou \textit{et al.}
%
%For the acoustic and visual modality, we concatenate the mean, max and min features on the %response-partitioned 
%video frames to achieve temporal aggregation as outlined by Zou \textit{et al.}.
%
\begin{comment}
However, there were no %provided 
ready-to-use feature vectors in E-DAIC unlike %the vectors provided by Zou et al. 
in CMDC. 
%
As a result, feature vectors for every modality (acoustic, textual, and visual) had to be extracted for every participant. 
%
For the fairest comparison between CMDC and E-DAIC, we generated these feature vectors in a similar fashion as done by Zou \textit{et al.}.
\end{comment}
%
%For the acoustic features, we use the OpenSMILE (version 3.0) \cite{eyben_OPENSMILE_2010} to extract the eGeMAPS features \cite{eyben_eGeMAPS_2016} in alignment with \cite{zou_cmdc_2023}.
%to Zou et al. \cite{zou_cmdc_2023}. 
%
%\begin{comment}
Across the visual and acoustic features, we therefore 
%Zou et al. mention evaluating their baseline methods on DAIC by 
concatenate the mean, max, and min features on the response-partitioned video frames to achieve temporal aggregation using the provided extracted features in the E-DAIC dataset. 
In E-DAIC, the same set of visual features and acoustic features were obtained using the OpenFace \cite{Baltrusaitis_OpenFace_2018} and openSMILE \cite{eyben_OPENSMILE_2010} toolkits respectively.
Readers can refer to \cite{ringeval_AVEC_2019,zou_cmdc_2023} for further details.
%
%The same methodology was used for temporal aggregation across acoustic and visual features per participant. 
%
%\end{comment}
%For generating the 
Across the textual modality, participant responses obtained via the provided transcripts were inputs to a pretrained English BERT model \cite{wolf-etal-2020-transformers} which resulted in a 768 feature long textual feature vector per participant.
%

%%%%%%%%%%%%%%%%%%%%%%%%%%%%%%%%%%%%%%%%%%%%%%%%
\subsection{ML Models and Training Details}

% -----------
\subsubsection{ML Models}
We investigate a binary classification setup where the models' goal is to predict whether a participant is classified as depressed or not. 
%
%A variety of popular ML classifier models were chosen for implementation. 
% 
In addition to the models introduced in \cite{zou_cmdc_2023}, we ran our experiments on more models to ensure thorough evaluation.
These models are the Support Vector Machine with a linear kernel (SVM Linear), Support Vector Machine with a radial basis function kernel (SVM RBF), Logistic Regression (LogReg), Naive Bayes (NB), K-Nearest Neighbors (KNN), Decision Tree (Dec Tree) and the Multilayer Perceptron (MLP). All models were implemented in Python using the sklearn package.
We have selected these basic models in alignment with prior works (e.g.,\cite{mathur2021modeling}, \cite{abbasi2022computational}, \cite{Cheong_NotFair_2023}) which employed statistical representations of features for training data as we have done in this work.
%
%All models were implemented in Python using the sklearn package, where sklearn's default values for each type of model were also used \cite{scikit-learn} to facilitate reproducibility.

% -----------
\subsubsection{Statistical Analysis}
We perform one-way multivariate ANOVA (MANOVA) tests across both the visual and acoustic features for both datasets.
%each question the participants faced in the semi-structured interview. 
%
%CMDC only provided textual features in the form of BERT embeddings whereas E-DAIC did not provide any textual features. 
%
\begin{comment}

%Thus, it was not feasible to perform further statistical analysis on these textual features using traditional MANOVA tests. 
%
%We omitted the textual features for both datasets as they were not provided in a feasible form t
%
%Section \ref{subsubsec:MANOVAVisualCMDC} covers the one-way MANOVA of visual features in CMDC, and section \ref{subsubsec:MANOVAAcousticCMDC} covers the one-way MANOVA of acoustic features in CMDC.
%
\end{comment}
%In order to address \textbf{RQ1}, 
%Since this research project also aims to 
%
%We also evaluated gender fairness across different ML models. 
%
We 
%accompany the one-way MANOVA with 
further conduct two-way MANOVA tests to compare how significantly different the means of visual and acoustic features are between participants within both the Healthy Control (HC) versus the Major Depressive Disorder (MDD) groups and the male versus female groups.
%
%The same one-way and two-way MANOVA statistical analysis employed on CMDC (see section \ref{subsec:CMDCStatisticalAnalysis}) was also performed on the visual and acoustic features provided in E-DAIC. Note that textual features were not provided in E-DAIC, only text transcripts were provided. Therefore it was not feasible to perform statistical analysis on provided textual data, as textual features could only be generated with methods outwith the provided dataset. When considering which frames to use for analysis, it was decided to remove frames outwith the time windows of participant responses provided in the transcripts. This was done to achieve a fairer comparison with CMDC's response-focused features.
%
All MANOVA analysis was implemented %and conducted in Python 
using the `statsmodels' package in Python \cite{seabold2010statsmodels}.
%
%In addition, 
Recent work indicates that analysing interviewees' responses at a question level allows for finer grained understanding and detection of behavioral characteristics of depression
\cite{guohou2020reveals}.
Thus we conduct the analysis on the CMDC dataset structured according to the questions posed within their semi-structured interview \cite{zou_cmdc_2023}.
%\cite{nordgaard2013psychiatric}

% -----------
\subsubsection{Training Process}
%Due to the limited number of participants
%
%
%and inherent class and gender imbalance present in both datasets 
%(see Tables \ref{tab:CMDCOverviewOfParticipants} \& \ref{tab:EDAICOverviewOfParticipants}), 
%the 
All models were evaluated using stratified 5-fold cross validation. 
We form stratified folds 
of equal ratios of depressed and non-depressed participants and also ensured equal ratios of male and female participants. 
This is to ensure that any observed disparities in model fairness across genders are reflective of the model's behavior rather than artefacts of the sampling process. 
The %defined 
folds are consistent across all classification tasks and models.
%performed with every model on each of the datasets. 
%
%\subsubsection{Uni vs Multiple Modalities}
%Another important factor chosen for evaluation was the effect that different feature modalities would have on each model's performance and gender fairness. 
%
Given that both datasets have acoustic (A), textual (T), and visual (V) features, the models were trained with every possible combination of the feature modalities (A, T, V, A+T, A+V, T+V, A+T+V).
%

%%%%%%%%%%%%%%%%%%%%%%%%%%%%%%%%%%%%%%%%%%%%%%%%
\subsection{Prediction and Fairness Measures}
We adopt the evaluation metrics overall accuracy,
%(Overall Acc), 
F1 Score and Area Under the Receiver Operating Characteristics (AUROC) used in \cite{zou_cmdc_2023} to evaluate each model's performance. 
$A=1$ denotes the majority group (male) and $A=0$ denotes the minority group (female).
$\hat{Y}$ denotes the predicted class.
%
%The chosen evaluation metrics for each model's fairness metrics across gender were 
We use Equal Accuracy ($EA_{Gender}$) and Disparate Impact ($DI_{Gender}$) to evaluate each model's fairness in alignment with existing works \cite{shen_mitigate_2020, zanna_bias_2022, Cheong_NotFair_2023}. 

\begin{itemize}
    \item \textbf{Equal Accuracy ($EA$)}, can be understood as the accuracy gap between the majority and the minority group: 
    %%%
    \begin{equation}
    \label{eqn:equal_accuracy}
    \ EA_{Gender} = |MAE (\hat{Y}| A=1) - MAE (\hat{Y}| A=0)|  ,
    \end{equation}
    %%%
    where $MAE$ represents the Mean Absolute Error (MAE) of the classification task of each sensitive group.
    \item \textbf{Disparate Impact ($DI$)}, measures the ratio of positive outcome ($\hat{Y}=1$) for the majority and minority groups as follows:
    %%%
    \begin{equation}
    \label{eqn:disparate_impact}
    \ DI_{Gender} = \frac{Pr(\hat{Y}=1|A=0 ) }{ Pr(\hat{Y}=1 | A=1)}
    \end{equation}
    %%% 
%    
\end{itemize}

$EA$ is dependent on predicted and actual outcome whereas $DI$ is mainly dependent on the predicted outcome. The complementary nature of both fairness measures thus provides a better understanding of the bias present.

%We employ the same prediction and fairness measures across all ML model experimentation and datasets.

%%%%%%%%%%%%%%%%%%%%%%
% PREPROCESSING DATA FROM DATASETS

%\section{Analysing Features Within the CMDC and E–-DAIC Datasets}

\section{Statistical Analysis Results}

%The 
%Our first contribution 
%of this paper 
%is to 
We evaluate whether there are any differences in depression manifestation across gender and countries. 
% We do so by presenting a statistical analysis of the provided visual, acoustic, and textual features present within the CMDC and E-DAIC datasets.
%
All statistical analysis is conducted at a significance value of $0.05$. 

% -----------

% -----------

\subsection{Statistical Analysis of Features in the CMDC Dataset}
\label{subsec:CMDCStatisticalAnalysis}
For CMDC, our statistical analysis 
% experimentation 
revealed significant differences between the different features representations across 
depressed vs. non-depressed classes 
as well as gender. 
%
%Further details on the results are presented in the subsections below.

%When considering what statistical analysis to perform on features available in the CMDC dataset, it became clear that the most relevant insight would come from comparing how significantly different the means of features are between participants in the HC and MDD groups. 
%
%Therefore, it was decided to 

%%We performed the one-way multivariate ANOVA (MANOVA) tests across both the provided visual and acoustic features for each question the participants faced in the semi-structured interview. 
%
%%Note that textual features were only provided in the form of BERT vector embeddings, therefore it was not feasible to perform further statistical analysis on these embeddings using traditional MANOVA tests. 
%%Section \ref{subsubsec:MANOVAVisualCMDC} covers the one-way MANOVA of visual features in CMDC, and section \ref{subsubsec:MANOVAAcousticCMDC} covers the one-way MANOVA of acoustic features in CMDC.

%In order to address \textbf{RQ1}, 
%Since this research project also aims to 
%%We also evaluated gender fairness across different ML models. We accompanied the one-way MANOVA with further two-way MANOVA tests to compare how significantly different the means of visual and acoustic features are between participants within both the HC/MDD groups and the male/female gender groups.

%All MANOVA analysis was implemented and conducted in Python using the `statsmodels' package \cite{seabold2010statsmodels}.

% --

\subsubsection{One-Way MANOVA (Visual)}
\label{subsubsec:MANOVAVisualCMDC}
%
%For the 45 participants who agreed to be video recorded, CMDC provides the OpenFace visual features \cite{Baltrusaitis_OpenFace_2018} for every frame of video recorded for each participant during each question. The means of the facial action unit, pose, and gaze angle features were taken for temporal aggregation, resulting in a single mean value for each participant's visual features per question.
%
%Table \ref{tab:CMDC_ONEWAYANDTWOWAYMANOVA_Visual_Data} shows the depression status one-way MANOVA analysis on the mean facial action unit classification visual features (FAUs), the mean pose visual features (Pose), and the mean gaze visual features (Gaze).
%
%%%%%%%
% COMBINING TABLES 6 & 7

\begin{table*}
  \begin{center}
  \caption{
  %\textcolor{magenta}
  {One-Way MANOVA (Depression Status)} \& %\textcolor{teal}
  {Two-Way MANOVA (Depression Status \& Gender)} for Visual Features in the CMDC Dataset. Abbreviation: FAUs: Facial Action Units.}
  \label{tab:CMDC_ONEWAYANDTWOWAYMANOVA_Visual_Data}
  \begin{tabular}{c|ccc|ccc}
    \toprule

     &\multicolumn{3}{c|}{One-Way MANOVA} &\multicolumn{3}{c}{Two-Way MANOVA } \\
     \midrule
    \textbf{Q No.} & \textbf{
    %\textcolor{magenta}
    {FAUs F(18,26) p}} & \textbf{
    %\textcolor{magenta}
    {Gaze F(8,36) p}} & \textbf{
    %\textcolor{magenta}
    {Pose F(6,38) p}} & \textbf{
    %\textcolor{teal}
    {FAUs F(18,24) p}} & \textbf{
    %\textcolor{teal}
    {Gaze F(8,34) p}} & \textbf{
    %\textcolor{teal}
    {Pose F(6,36) p}}\\
    \midrule
    Q1 & \textbf{0.01} & 0.46 & \textbf{$<$ 0.00} & 0.26 & 0.31 & 0.21\\
    Q3 & \textbf{0.02} & 0.26 & \textbf{$<$ 0.00} & 0.56 & 0.93 & 0.71\\
    Q4 & 0.06 & 0.56 & \textbf{$<$ 0.00} & 0.11 & 0.91 & 0.75\\
    Q5 & 0.24 & 0.80 & \textbf{$<$ 0.00} & 0.40 & 0.68 & 0.72\\
    Q6 & 0.42 & 0.82 & \textbf{$<$ 0.00} & 0.46 & 0.43 & 0.73\\
    Q7 & \textbf{0.02} & 0.91 & \textbf{$<$ 0.00} & 0.16 & 0.54 & 0.41\\
    Q9 & 0.28 & 0.38 & 0.11 & 0.54 & 0.61 & 0.40\\
  \bottomrule
\end{tabular}
  \vspace{-0.3cm}
\end{center}
\end{table*}
%%%%%%%
%

Looking at Table \ref{tab:CMDC_ONEWAYANDTWOWAYMANOVA_Visual_Data},
%and considering 
%at a significance value of 0.05, 
there are statistically significant differences between the HC/MDD groups for the Action Unit (AU) classification features for all questions except 4, 5, 6, and 9. 
There are statistically significant differences between the HC/MDD groups for the pose features for all questions except 9. Interestingly, 
%it is also clear that 
there are no statistically significant differences between the HC/MDD groups for the gaze features across all questions. 
%
%\textbf{Takeaway:}
This indicates that \textbf{participants labelled to have MDD} in CMDC indeed \textbf{exhibit different facial action unit activations and different head orientations (poses)} compared to HC participants, however they do not exhibit significantly different gaze directions.

% --

\subsubsection{Two-Way MANOVA (Visual)}
\label{subsubsec:TwoWayMANOVAVisualCMDC}
%Table \ref{tab:CMDC_ONEWAYANDTWOWAYMANOVA_Visual_Data} shows the depression status and gender two-way MANOVA analysis on the same visual features as specified in section \ref{subsubsec:MANOVAVisualCMDC}.
%
%From observing 
Table \ref{tab:CMDC_ONEWAYANDTWOWAYMANOVA_Visual_Data} 
%and considering a significance value of 0.05, 
indicates 
%%that there are 
no statistically significant differences between the HC/MDD and Male/Female groups for any of the OpenFace visual feature, % subsets. 
%\textbf{Takeaway:}
%This 
indicating that the \textbf{effect of depression status on the visual features does not vary significantly by gender}.

% --

\subsubsection{One-Way MANOVA (Acoustic)}
\label{subsubsec:MANOVAAcousticCMDC}
%CMDC provides the eGeMAPS features \cite{eyben_eGeMAPS_2016} extracted by OpenSmile \cite{eyben_OPENSMILE_2010} for each participant during each question. Zou et al. specify that they used 8 frequency-related parameters (Freq), 3 amplitude related parameters (Amp),  and 14 spectral parameters (Spec) \cite{zou_cmdc_2023} and so these same features are used. The means of these frequency, amplitude, and spectral features were taken for temporal aggregation, and this resulted in a mean acoustic feature vector for every participant.
%
%Table \ref{tab:CMDC_ONEWAYANDTWOWAYMANOVA_Acoustic_Data} shows the depression status one-way MANOVA analysis on the mean acoustic features subgroups per question.
%
%%%%%%%%%%%%
% COMBINING TABLES 8 & 9
\begin{table*}
\begin{center}

  \caption{
  %\textcolor{magenta}
  {One-Way MANOVA (Depression Status)} \& 
  %\textcolor{teal}
  {Two-Way MANOVA (Depression Status \& Gender)} for Acoustic Features in the CMDC Dataset. Abbreviation: FAUs: Facial Action Units.}
  \label{tab:CMDC_ONEWAYANDTWOWAYMANOVA_Acoustic_Data}
  \begin{tabular}{c|ccc|ccc}
    \toprule
     &\multicolumn{3}{c|}{One-Way MANOVA} &\multicolumn{3}{c}{Two-Way MANOVA } \\
     \midrule
    \textbf{Q No.} & \textbf{
    %\textcolor{magenta}
    {Freq F(8,69) p}} & \textbf{
    %\textcolor{magenta}
    {Amp F(3,74) p}} & \textbf{
    %\textcolor{magenta}
    {Spec F(14,63) p}} & \textbf{
    %\textcolor{teal}
    {Freq F(8,67) p}} & \textbf{
    %\textcolor{teal}
    {Amp F(3,72) p}} & \textbf{
    %\textcolor{teal}
    {Spec F(14,61) p}}\\
    \midrule
    Q1 & \textbf{$<$ 0.00} & \textbf{$<$ 0.00} & \textbf{$<$ 0.00} & 0.10 & 0.05 & \textbf{0.02}\\
    Q2 & \textbf{$<$ 0.00} & \textbf{$<$ 0.00} & \textbf{$<$ 0.00} & 0.35 & \textbf{0.01} & \textbf{$<$ 0.00}\\
    Q3 & \textbf{$<$ 0.00} & \textbf{$<$ 0.00} & \textbf{$<$ 0.00} & 0.88 & 0.10 & \textbf{0.04}\\
    Q4 & \textbf{$<$ 0.00} & \textbf{$<$ 0.00} & \textbf{$<$ 0.00} & 0.73 & 0.11 & \textbf{$<$ 0.00}\\
    Q6 & \textbf{$<$ 0.00} & \textbf{$<$ 0.00} & \textbf{$<$ 0.00} & 0.14 & \textbf{0.02} & \textbf{$<$ 0.01}\\
    Q7 & \textbf{$<$ 0.00} & \textbf{$<$ 0.00} & \textbf{$<$ 0.00} & \textbf{0.02} & 0.10 & \textbf{$<$ 0.01}\\
    Q9 & \textbf{$<$ 0.00} & \textbf{$<$ 0.00} & \textbf{$<$ 0.00} & 0.06 & \textbf{0.03} & \textbf{0.01}\\
    Q10 & \textbf{$<$ 0.00} & \textbf{$<$ 0.00} & \textbf{$<$ 0.00} & 0.05 & \textbf{0.02} & \textbf{$<$ 0.01}\\
    Q11 & \textbf{$<$ 0.00} & \textbf{$<$ 0.00} & \textbf{$<$ 0.00} & 0.08 & \textbf{0.01} & \textbf{0.01}\\
    Q12 & \textbf{$<$ 0.00} & \textbf{$<$ 0.00} & \textbf{$<$ 0.00} & 0.27 & \textbf{0.03} & \textbf{0.02}\\
  \bottomrule
\end{tabular}
\end{center}
  \vspace{-0.2cm}
\end{table*}
%%%%%%%%%%%%
%
\begin{table*}
\begin{center}
  \caption{
  %\textcolor{magenta}
  {One-Way MANOVA (Depression Status)} \& 
  %\textcolor{teal}
  {Two-Way MANOVA (Depression Status \& Gender)} for Visual Features in the E-DAIC Dataset. Abbreviation: FAUs: Facial Action Units.}
  \label{tab:EDAIC_ONEWAYANDTWOWAYMANOVA_Visual_Data}
  \begin{tabular}{ccc|ccc}
    \toprule
     \multicolumn{3}{c|}{One-Way MANOVA} &\multicolumn{3}{c}{Two-Way MANOVA} \\
     \midrule
    \textbf{%\textcolor{magenta}
    {FAUs F(18,256) p}} & \textbf{%\textcolor{magenta}
    {Gaze F(8,266) p}} & \textbf{%\textcolor{magenta}
    {Pose F(6,268) p}} & \textbf{%\textcolor{teal}
    {FAUs F(18,254) p}} & \textbf{%\textcolor{teal}
    {Gaze F(8,264) p}} & \textbf{%\textcolor{teal}
    {Pose F(6,266) p}}\\
    \midrule
    0.57 & \textbf{0.04} & 0.19 & 0.54 & 0.45 & 0.31\\
  \bottomrule
\end{tabular}
  \vspace{-0.2cm}
\end{center}
\end{table*}
%
%
% MAKING A TABLE OF E-DAIC STATISTICAL ANALYSIS ON ACOUSTIC FEATURES
\begin{table*}
\begin{center}
  \caption{%\textcolor{magenta}
  {One-Way MANOVA (Depression Status)} \& %\textcolor{teal}
  {Two-Way MANOVA (Depression Status \& Gender)} for Acoustic Features in the E-DAIC Dataset. Abbreviation: FAUs: Facial Action Units.}
  \label{tab:EDAIC_ONEWAYANDTWOWAYMANOVA_Acoustic_Data}
  \begin{tabular}{ccc|ccc}
    \toprule
     \multicolumn{3}{c|}{One-Way MANOVA} &\multicolumn{3}{c}{Two-Way MANOVA} \\
     \midrule
    \textbf{%\textcolor{magenta}
    {Freq F(6,268) p}} & \textbf{%\textcolor{magenta}
    {Amp F(3,271) p}} & \textbf{%\textcolor{magenta}
    {Spec F(14,260) p}} & \textbf{%\textcolor{teal}
    {Freq F(6,266) p}} & \textbf{%\textcolor{teal}
    {Amp F(3,269) p}} & \textbf{%\textcolor{teal}
    {Spec F(14,258) p}}\\
    \midrule
    0.24 & 0.17 & 0.34 & 0.21 & 0.41 & 0.32\\
  \bottomrule
\end{tabular}
\end{center}
\end{table*}
%
%%%%%%%%%%%%%%%%%%%%%%

%\begin{comment}

\begin{table*}[ht]
\begin{center}

  \caption{Model Peformance \& Gender Fairness Results on the CMDC Dataset (the best values for each metric are in \textbf{{bold}}, and the worst values for each metric are \underline{{underlined}})}
  \label{tab:RESULTSCMDC}
  \begin{tabular}{c|lccccccc}
    \toprule
    \textbf{Features} & \textbf{Metric} & \textbf{SVM Linear} & \textbf{SVM RBF} & \textbf{LogReg} & \textbf{Naive Bayes} & \textbf{KNN} & \textbf{Dec Tree} & \textbf{MLP}\\
    \midrule
    \multirow{5}{*}{A} & Overall Acc   & \textbf{{0.94}} & 0.91 & \textbf{{0.94}} & \underline{{0.86}} & 0.91 & 0.90 & 0.93\\
                       & F1 Score      & \textbf{{0.92}} & 0.85 & \textbf{{0.92}} & \underline{{0.80}} & 0.84 & 0.84 & 0.90\\
                       & AUROC         & 0.96 & 0.95 & \textbf{{0.97}} & \underline{{0.86}} & 0.94 & 0.91 & 0.93\\
                       \cmidrule{2-9}
                       & $EA_{Gender}$ & 0.03 & \textbf{{0.01}} & 0.03 & \underline{{0.09}} & \textbf{{0.01}} & 0.03 & 0.07\\
                       & $DI_{Gender}$ & 0.51 & 0.56 & 0.51 & \underline{{0.44}} & \textbf{{0.75}} & 0.69 & 0.69\\
    \midrule
    \multirow{5}{*}{T} & Overall Acc   & \textbf{{0.88}} & 0.81 & \textbf{{0.88}} & 0.75 & 0.78 & \underline{{0.65}} & 0.79\\
                       & F1 Score      & \textbf{{0.79}} & 0.66 & \textbf{{0.79}} & 0.60 & \underline{{0.36}} & 0.44 & 0.68\\
                       & AUROC         & \textbf{{0.91}} & 0.84 & \textbf{{0.91}} & 0.81 & 0.72 & \underline{{0.60}} & 0.88\\
                       \cmidrule{2-9}
                       & $EA_{Gender}$ & \textbf{{0.00}} & 0.10 & \textbf{{0.00}} & 0.07 & \underline{{0.15}} & 0.07 & 0.10\\
                       & $DI_{Gender}$ & 1.07 & \textbf{{1.00}} & 1.07 & 0.92 & \underline{{2.40}} & 0.90 & 1.10\\
    \midrule
    \multirow{5}{*}{V} & Overall Acc   & 0.96 & 0.93 & 0.96 & 0.93 & \underline{{0.89}} & 0.91 & \textbf{{0.98}}\\
                       & F1 Score      & 0.92 & 0.89 & 0.92 & 0.89 & \underline{{0.78}} & 0.87 & \textbf{{0.96}}\\
                       & AUROC         & \textbf{{1.00}} & 0.98 & 0.99 & 0.94 & \underline{{0.92}} & \underline{{0.92}} & 0.98\\
                       \cmidrule{2-9}
                       & $EA_{Gender}$ & \underline{{0.13}} & 0.10 & \underline{{0.13}} & 0.10 & \underline{{0.13}} & \textbf{{0.03}} & 0.07\\
                       & $DI_{Gender}$ & \textbf{{1.00}} & \textbf{{1.00}} & \textbf{{1.00}} & \textbf{{1.00}} & \underline{{1.50}} & 1.27 & \textbf{{1.00}}\\
    \midrule
    \multirow{5}{*}{A+T} & Overall Acc & \textbf{{0.98}} & 0.95 & \textbf{{0.98}} & \underline{{0.85}} & 0.88 & 0.88 & 0.93\\
                       & F1 Score      & \textbf{{0.96}} & 0.91 & \textbf{{0.96}} & 0.77 & \underline{{0.73}} & 0.84 & 0.90\\
                       & AUROC         & \textbf{{0.98}} & \textbf{{0.98}} & \textbf{{0.98}} & \underline{{0.86}} & 0.93 & 0.91 & 0.97\\
                       \cmidrule{2-9}
                       & $EA_{Gender}$ & 0.02 & \textbf{{0.01}} & 0.02 & \underline{{0.16}} & 0.05 & 0.06 & \textbf{{0.01}}\\
                       & $DI_{Gender}$ & 0.57 & 0.54 & 0.57 & \underline{{0.43}} & \textbf{{0.93}} & 0.75 & 0.66\\
    \midrule
    \multirow{5}{*}{A+V} & Overall Acc & \textbf{{0.98}} & 0.93 & \textbf{{0.98}} & \underline{{0.91}} & \textbf{{0.98}} & 0.93 & \textbf{{0.98}}\\
                       & F1 Score      & \textbf{{0.96}} & 0.90 & \textbf{{0.96}} & \underline{{0.85}} & \textbf{{0.96}} & 0.89 & \textbf{{0.96}}\\
                       & AUROC         & \textbf{{1.00}} & 0.99 & \textbf{{1.00}} & \underline{{0.88}} & 0.97 & 0.92 & 0.99\\
                       \cmidrule{2-9}
                       & $EA_{Gender}$ & 0.07 & \textbf{{0.00}} & 0.07 & 0.07 & 0.07 & \underline{{0.20}} & 0.07\\
                       & $DI_{Gender}$ & \textbf{{1.00}} & 0.83 & \textbf{{1.00}} & \underline{{1.33}} & \textbf{{1.00}} & \underline{{1.33}} & \textbf{{1.00}}\\
    \midrule
    \multirow{5}{*}{T+V} & Overall Acc & \textbf{{0.98}} & 0.96 & \textbf{{0.98}} & 0.93 & \underline{{0.89}} & \underline{{0.89}} & 0.93\\
                       & F1 Score      & \textbf{{0.96}} & 0.93 & \textbf{{0.96}} & 0.91 & \underline{{0.78}} & 0.82 & 0.91\\
                       & AUROC         & \textbf{{1.00}} & 0.96 & \textbf{{1.00}} & 0.93 & 0.92 & \underline{{0.87}} & 0.98\\
                       \cmidrule{2-9}
                       & $EA_{Gender}$ & 0.07 & 0.03 & 0.07 & \textbf{{0.00}} & \underline{{0.13}} & 0.03 & \textbf{{0.00}}\\
                       & $DI_{Gender}$ & \textbf{{1.00}} & 0.92 & \textbf{{1.00}} & 0.88 & 1.50 & \underline{{1.67}} & 0.88\\
    \midrule
    \multirow{5}{*}{A+T+V} & Overall Acc & \textbf{{0.98}} & 0.93 & \textbf{{0.98}} & \underline{{0.91}} & 0.93 & \underline{{0.91}} & 0.96\\
                       & F1 Score        & \textbf{{0.96}} & 0.90 & \textbf{{0.96}} & \underline{{0.85}} & 0.88 & 0.86 & 0.93\\
                       & AUROC           & \textbf{{1.00}} & 0.99 & \textbf{{1.00}} & \underline{{0.88}} & 0.97 & 0.90 & 0.99\\
                       \cmidrule{2-9}
                       & $EA_{Gender}$   & 0.07 & \textbf{{0.00}} & 0.07 & 0.07 & \underline{{0.20}} & 0.17 & 0.03\\
                       & $DI_{Gender}$   & \textbf{{1.00}} & 0.83 & \textbf{{1.00}} & \underline{{1.33}} & \textbf{{1.00}} & 1.22 & 0.92\\
    \midrule
  \bottomrule
\end{tabular}
\end{center}
\end{table*}
%%%%%%%%%%%%%%%%%%%%%%
%\end{comment}

%%%%%%%%%%%%%%%%%%%%%%
%\begin{comment}

\begin{table*}[ht]
\begin{center}
  \caption{Model Peformance \& Gender Fairness Results on the E-DAIC Dataset (the best values for each metric are in \textbf{{bold}}, and the worst values for each metric are \underline{{underlined}})}
  \label{tab:RESULTSEDAIC}
  \begin{tabular}{c|lccccccc}
    \toprule
    \textbf{Features} & \textbf{Metric} & \textbf{SVM Linear} & \textbf{SVM RBF} & \textbf{LogReg} & \textbf{Naive Bayes} & \textbf{KNN} & \textbf{Dec Tree} & \textbf{MLP}\\
    \midrule
    \multirow{5}{*}{A} & Overall Acc   & 0.68 & 0.60 & 0.67 & 0.65 & \underline{{0.59}} & 0.66 & \textbf{{0.69}}\\
                       & F1 Score      & \textbf{{0.80}} & 0.72 & 0.79 & 0.77 & \underline{{0.71}} & 0.77 & \textbf{{0.80}}\\
                       & AUROC         & 0.70 & 0.65 & 0.69 & 0.68 & \underline{{0.63}} & 0.68 & \textbf{{0.72}}\\
                       \cmidrule{2-9}
                       & $EA_{Gender}$ & 0.11 & 0.12 & \underline{{0.15}} & 0.08 & 0.13 & 0.06 & \textbf{{0.05}}\\
                       & $DI_{Gender}$ & 0.98 & \underline{{0.81}} & 0.93 & 0.99 & 0.95 & 0.92 & \textbf{{1.00}}\\
    \midrule
    \multirow{5}{*}{T} & Overall Acc   & 0.67 & 0.67 & \textbf{{0.70}} & \underline{{0.51}} & 0.52 & 0.65 & \textbf{{0.70}}\\
                       & F1 Score      & 0.78 & 0.77 & 0.80 & \underline{{0.58}} & 0.62 & 0.77 & \textbf{{0.81}}\\
                       & AUROC         & 0.70 & \textbf{{0.75}} & 0.73 & 0.70 & \underline{{0.60}} & 0.72 & 0.73\\
                       \cmidrule{2-9}
                       & $EA_{Gender}$ & 0.02 & 0.03 & \textbf{{0.01}} & \underline{{0.12}} & 0.03 & 0.09 & 0.02\\
                       & $DI_{Gender}$ & 1.10 & 1.10 & 1.11 & \underline{{1.76}} & 1.15 & \textbf{{1.06}} & 1.08\\
    \midrule
    \multirow{5}{*}{V} & Overall Acc   & \textbf{{0.67}} & 0.57 & 0.64 & \underline{{0.53}} & 0.57 & 0.65 & 0.61\\
                       & F1 Score      & \textbf{{0.78}} & 0.70 & 0.77 & \underline{{0.53}} & 0.68 & \textbf{{0.78}} & 0.74\\
                       & AUROC         & \textbf{{0.72}} & 0.60 & 0.66 & \underline{{0.54}} & 0.63 & 0.69 & 0.65\\
                       \cmidrule{2-9}
                       & $EA_{Gender}$ & 0.01 & 0.01 & 0.05 & \textbf{{0.00}} & 0.05 & \underline{{0.07}} & 0.02\\
                       & $DI_{Gender}$ & 1.08 & 1.15 & 1.16 & \underline{{0.82}} & 1.13 & \textbf{{1.05}} & 1.14\\
    \midrule
    \multirow{5}{*}{A+T} & Overall Acc & 0.63 & 0.66 & 0.65 & \underline{{0.52}} & \underline{{0.52}} & 0.61 & \textbf{{0.68}}\\
                       & F1 Score      & 0.75 & 0.77 & 0.77 & \underline{{0.60}} & 0.62 & 0.73 & \textbf{{0.80}}\\
                       & AUROC         & 0.67 & 0.69 & 0.70 & 0.58 & \underline{{0.57}} & 0.68 & \textbf{{0.74}}\\
                       \cmidrule{2-9}
                       & $EA_{Gender}$ & 0.02 & 0.03 & \textbf{{0.00}} & \underline{{0.05}} & 0.02 & 0.04 & 0.04\\
                       & $DI_{Gender}$ & 1.06 & 1.02 & 1.03 & \underline{{1.58}} & 1.15 & 1.03 & \textbf{{0.99}}\\
    \midrule
    \multirow{5}{*}{A+V} & Overall Acc & 0.54 & 0.61 & 0.60 & \underline{{0.53}} & 0.55 & 0.57 & \textbf{{0.62}}\\
                       & F1 Score      & 0.66 & 0.74 & 0.73 & \underline{{0.55}} & 0.60 & 0.71 & \textbf{{0.75}}\\
                       & AUROC         & 0.64 & 0.69 & 0.70 & \underline{{0.53}} & 0.58 & 0.70 & \textbf{{0.71}}\\
                       \cmidrule{2-9}
                       & $EA_{Gender}$ & 0.02 & \underline{{0.12}} & \textbf{{0.01}} & \textbf{{0.01}} & 0.09 & 0.08 & 0.03\\
                       & $DI_{Gender}$ & 1.04 & \underline{{0.85}} & 1.06 & 0.90 & \textbf{{0.99}} & 0.97 & 1.02\\
    \midrule
    \multirow{5}{*}{T+V} & Overall Acc & 0.66 & 0.65 & \textbf{{0.68}} & \underline{{0.50}} & 0.52 & 0.63 & \textbf{{0.68}}\\
                       & F1 Score      & 0.78 & 0.76 & \textbf{{0.79}} & \underline{{0.56}} & 0.62 & 0.75 & \textbf{{0.79}}\\
                       & AUROC         & 0.72 & 0.70 & \textbf{{0.74}} & 0.58 & \underline{{0.55}} & 0.71 & 0.73\\
                       \cmidrule{2-9}
                       & $EA_{Gender}$ & 0.08 & \textbf{{0.01}} & 0.06 & \underline{{0.13}} & 0.03 & 0.04 & 0.05\\
                       & $DI_{Gender}$ & 1.11 & 1.04 & 1.08 & \underline{{1.72}} & 1.18 & \textbf{{1.02}} & 1.05\\
    \midrule
    \multirow{5}{*}{A+T+V} & Overall Acc & 0.64 & 0.65 & 0.66 & 0.52 & \underline{{0.51}} & 0.59 & \textbf{{0.68}}\\
                       & F1 Score        & 0.76 & 0.77 & 0.78 & \underline{{0.58}} & 0.61 & 0.72 & \textbf{{0.80}}\\
                       & AUROC           & 0.69 & 0.73 & 0.71 & \underline{{0.58}} & \underline{{0.58}} & 0.66 & \textbf{{0.75}}\\
                       \cmidrule{2-9}
                       & $EA_{Gender}$   & 0.05 & \textbf{{0.01}} & 0.10 & \underline{{0.12}} & 0.04 & 0.03 & 0.06\\
                       & $DI_{Gender}$   & 1.04 & \textbf{{1.00}} & 1.09 & \underline{{1.58}} & 0.96 & 1.04 & 1.03\\
    \midrule
  \bottomrule
\end{tabular}
\end{center}
\end{table*}
%\end{comment}

\begin{table*}[ht!]
\begin{center}
\addtolength{\tabcolsep}{-1mm}
%\small
  \caption{Best Model Peformance (MLP) and Gender Fairness Results on CMDC and E-DAIC (the best values for each metric are in \textbf{{bold}}).}
  \label{tab:BESTRESULTS}
  \begin{tabular}{c|ccccc|ccccc}
    \toprule
    & \multicolumn{5}{c|}{CMDC} &\multicolumn{5}{c}{E-DAIC} \\
    \midrule
    \textbf{Features} & \textbf{Overall Acc} & \textbf{F1 Score} & \textbf{AUROC} & \textbf{$EA_{Gender}$} & \textbf{$DI_{Gender}$} & \textbf{Overall Acc} & \textbf{F1 Score} & \textbf{AUROC} & \textbf{$EA_{Gender}$} & \textbf{$DI_{Gender}$}\\
    
    \midrule
   A & 0.93 & 0.90 & 0.93 & 0.07 & 0.69 & \textbf{0.69} & 0.80 & 0.72 & 0.05 & \textbf{1.00}\\
    \midrule
    T & 0.79 & 0.68 & 0.88 & 0.10 & 1.10 & 0.70 &\textbf{ 0.81} & 0.73 & \textbf{0.02} & 1.08\\ 
    \midrule
    V & \textbf{0.98} & \textbf{0.96} & 0.98 & 0.07 & \textbf{1.00} & 0.61 & 0.74 & 0.65 &\textbf{ 0.02} & 1.14 \\ 
    \midrule
    A+T & 0.93 & 0.90 & 0.97 & 0.01 & 0.66 & 0.68 & 0.80 & 0.74 & 0.04 & 0.99\\ 
    \midrule
    A+V & \textbf{0.98} & \textbf{0.96} &\textbf{ 0.99 }& 0.07 & \textbf{1.00} & 0.62 & 0.75 & 0.71& 0.03 & 1.02 \\ 
    \midrule
    T+V & 0.93 & 0.91 & 0.98 &\textbf{ 0.00 }& 0.88  & 0.68 & 0.79 & 0.73 & 0.05 & 1.05\\ 
    \midrule
    A+T+V & 0.96 & 0.93 & \textbf{0.99 }& 0.03 & 0.92 & 0.68& 0.80 & \textbf{0.75 }& 0.06 & 1.03\\ 
    \midrule
  \bottomrule
\end{tabular}
\end{center}
\end{table*}

\begin{comment}

\begin{table*}[ht]
\begin{center}

%\small
  \caption{Best Model Peformance (MLP) \& Gender Fairness Results on the E-DAIC Dataset (the best values for each metric are in \textbf{{bold}})}
  \label{tab:RESULTSEDAIC}
  \begin{tabular}{c|ccccc}
    \toprule
    \textbf{Features} & \textbf{Overall Acc} & \textbf{F1 Score} & \textbf{AUROC} & \textbf{$EA_{Gender}$} & \textbf{$DI_{Gender}$}\\
    \midrule
   A & \textbf{0.69} & 0.80 & 0.72 & 0.05 & \textbf{1.00} \\
    \midrule
    T & 0.70 &\textbf{ 0.81} & 0.73 & \textbf{0.02} & 1.08 \\ 
    \midrule
    V & 0.61 & 0.74 & 0.65 &\textbf{ 0.02} & 1.14 \\ 
    \midrule
    A+T & 0.68 & 0.80 & 0.74 & 0.04 & 0.99 \\ 
    \midrule
    A+V & 0.62 & 0.75 & 0.71& 0.03 & 1.02 \\ 
    \midrule
    T+V & 0.68 & 0.79 & 0.73 & 0.05 & 1.05 \\ 
    \midrule
    A+T+V & 0.68& 0.80 & \textbf{0.75 }& 0.06 & 1.03 \\ 
    \midrule
  \bottomrule
\end{tabular}
\end{center}
\end{table*}
%
\end{comment}
%

Table \ref{tab:CMDC_ONEWAYANDTWOWAYMANOVA_Acoustic_Data} %and considering a significance value of 0.05,
indicates that 
%there are statistically significant differences between the HC/MDD groups for all acoustic features across all questions.
%\textbf{Takeaway:}
%This indicates that 
\textbf{participants labelled to have MDD} in CMDC indeed \textbf{exhibit different pitch, amplitude, and spectral sound characteristics compared to HC participants}.

% --

\subsubsection{Two-Way MANOVA (Acoustic)}
\label{subsubsec:TwoWayMANOVAAcousticCMDC}
%Table \ref{tab:CMDC_ONEWAYANDTWOWAYMANOVA_Acoustic_Data} shows the depression status and gender two-way MANOVA analysis on the same acoustic features as specified in section \ref{subsubsec:MANOVAAcousticCMDC}.
%
With reference to Table \ref{tab:CMDC_ONEWAYANDTWOWAYMANOVA_Acoustic_Data} 
%and considering a significance value of 0.05, 
there are 
%indeed 
statistically significant differences between the HC/MDD and Male/Female groups for some of the OpenSMILE acoustic feature subsets. 
The spectral acoustic features are significantly different between the HC/MDD and Male/Female groups for all questions. The amplitude feature are 
%significantly 
different between the HC/MDD and Male/Female groups for some questions. 
%\textbf{Takeaway:}
This indicates that the effect of depression status (HC/MDD) on the acoustic features, especially the \textbf{spectral acoustic features and the amplitude acoustic features} for some questions, \textbf{does 
%indeed 
vary across gender}.

% -----------

%%%%%%%%%%%%%%%%%%%%%%%%%%%%%%%%%%%%%%%%%%%%%%%%%%%%%%%%%%%%

\subsection{Statistical Analysis of Features in the E-DAIC Dataset}
\label{subsec:EDAICStatisticalAnalysis}

For E-DAIC, the statistical analysis reveal that there are significant differences between the gaze features of depressed and non-depressed participants. 
%
%Further details on each statistical analysis conducted are presented in the subsections below.

%The same one-way and two-way MANOVA statistical analysis employed on CMDC (see section \ref{subsec:CMDCStatisticalAnalysis}) was also performed on the visual and acoustic features provided in E-DAIC. Note that textual features were not provided in E-DAIC, only text transcripts were provided. Therefore it was not feasible to perform statistical analysis on provided textual data, as textual features could only be generated with methods outwith the provided dataset. When considering which frames to use for analysis, it was decided to remove frames outwith the time windows of participant responses provided in the transcripts. This was done to achieve a fairer comparison with CMDC's response-focused features.

% --
%
% MAKING A TABLE OF E-DAIC STATISTICAL ANALYSIS ON VISUAL FEATURES

\subsubsection{One-Way MANOVA (Visual)}
\label{subsubsec:MANOVAVisualEDAIC}
%E-DAIC provides the OpenFace visual features \cite{Baltrusaitis_OpenFace_2018} for every frame of video recorded for each participant. The means of the classification facial action unit (FAUs), pose, and gaze features were taken for temporal aggregation per participant. 
%The depression status one-way MANOVA analysis for the visual features subgroups can be seen in 
Table \ref{tab:EDAIC_ONEWAYANDTWOWAYMANOVA_Visual_Data} indicates that
%
%For a significance level of 0.05, 
there are \textbf{significant differences between the gaze features} of depressed and non-depressed participants.

% --

\subsubsection{Two-Way MANOVA (Visual)}
\label{subsubsec:TwoWayMANOVAVisualEDAIC}
%The depression status and gender two-way MANOVA analysis on the same visual features can also be seen in Table \ref{tab:EDAIC_ONEWAYANDTWOWAYMANOVA_Visual_Data}. 
%
%With reference to 
Table \ref{tab:EDAIC_ONEWAYANDTWOWAYMANOVA_Visual_Data}
%, considering a significance level of 0.05, this analysis 
indicates that the effect of depression status on the visual features \textbf{does not vary significantly} by gender in E-DAIC.

% --

\subsubsection{One-Way MANOVA (Acoustic)}
\label{subsubsec:MANOVAAcousticEDAIC}
%
%E-DAIC provides the eGeMAPS features \cite{eyben_eGeMAPS_2016} extracted by OpenSmile \cite{eyben_OPENSMILE_2010} per video frame per participant. The means of the same acoustic features subgroups as described in section \ref{subsubsec:MANOVAAcousticCMDC} were taken for temporal aggregation per participant. 
%
%The depression status one-way MANOVA analysis for the acoustic features can be seen in 
Table \ref{tab:EDAIC_ONEWAYANDTWOWAYMANOVA_Acoustic_Data}
%. For a significance level of 0.05, 
indicates 
%that there are 
\textbf{no significant differences between the acoustic features} of depressed and non-depressed participants.

% --

\subsubsection{Two-Way MANOVA (Acoustic)}
\label{subsubsec:TwoWayMANOVAAcousticEDAIC}
%The depression and gender two-way MANOVA analysis on the acoustic features is captured in 
%
Table \ref{tab:EDAIC_ONEWAYANDTWOWAYMANOVA_Acoustic_Data}
%. For a significance level of 0.05, this 
indicates that the effect of depression status on the acoustic features \textbf{does not vary significantly by gender} in E-DAIC.

%%%%%%%%%%%%%%%%%%%%%%
%\subsection{Findings on RQ 1:}
%\textcolor{red}{WIP}

%%%%%%%%%%%%%%%%%%%%%%
%\section{Classification Modelling and Fairness Results on the CMDC \& E-DAIC Datasets}

\section{Experimental Results}

In this section, we evaluate the performance and fairness across all the different ML models for both datasets.
% ------
\subsection{Evaluating Models on the CMDC Dataset}

%Zou et al. provided ready-to-use feature vectors for every feature modality in the CMDC dataset and recommended using them for future machine learning tasks. Hence, these feature vectors were used in our current binary depression classification experimental setup. 
%
%As discussed in Section \ref{subsubsec:CMDCMissingData} however, there's missing data in the CMDC dataset. The same data that was cut from the acoustic, textual, and visual features for performing statistical analysis in section \ref{subsec:CMDCStatisticalAnalysis} was also cut for the classification tasks on the CMDC dataset. 
%
%We pre-processed the data as discussed in Section \ref{subsubsec:CMDCMissingData}.
%
%For any feature combination containing visual features, the stratified k-folds were based on the reduced sample size of participants who agreed to be video recorded. These considerations ensured that all data being used to train and test the classifiers contained data from certain questions appearing in the same locations within the input feature vector across all valid participants for specified feature combinations.
%
%Following these considerations for the CMDC dataset, the classification tasks were performed and 
%
%The results for CMDC are presented in Table \ref{tab:RESULTSCMDC}.

\subsubsection{Performance Evaluation}
%
%\textcolor{teal}{
With reference to Table \ref{tab:RESULTSCMDC}, 
%we see that 
the SVM Linear and Logistic Regression models consistently provide the best results across all performance measures. 
Both models produce the best accuracy, F1, and AUROC scores for the textual and all multimodal approaches.
Most models perform better within a multimodal setting. 
NB performs the worst across most performance measures.
For instance, across the audio modality, NB gave the lowest accuracy, F1, and AUROC scores of $0.86$, $0.80$ and $0.86$ respectively as well as the lowest multimodal (A+T+V) results of $0.91$, $0.85$ and $0.88$ across accuracy, F1 and AUROC as well.
This is followed by the KNN and Decision Tree models thus indicating these models may not be suited for the task or dataset at hand.
%}

\subsubsection{Fairness Evaluation}

%Similar to our performance evaluation above, 
Across the fairness measures, 
%results indicate that 
most models provide relatively good results. 
We see from Table \ref{tab:RESULTSCMDC} that the SVM Linear, SVM RBF and Logistic Regression models all provide consistently good results across most features and modalities.
For instance, both models produce the fairest $EA_{Gender}$ score of $0.00$ across the textual modality and the fairest $DI_{Gender}$ score of $1.00$ across the visual, A+V, T+V, and A+T+V modalities.
NB consistently produces the least fair outcome across most features and modalities followed by the KNN and Decision Tree models.

% ------
\subsection{Evaluating Models on the E-DAIC Dataset}

%As noted 
%during the statistical analysis phase (see section \ref{subsec:EDAICStatisticalAnalysis}), 
%in Section \ref{subsec:EDAICStatisticalAnalysis}, there was no missing data in the E-DAIC dataset. 
%
%We adopt the data pre-processing methods highlighted in Section \ref{sect:edaic_preprocessing}.
%
%Following these considerations for the E-DAIC dataset, the classification tasks were performed 
%The results for EDAIC are presented in Table \ref{tab:RESULTSEDAIC}.

\subsubsection{Performance Evaluation}
With reference to Table \ref{tab:RESULTSEDAIC}, the best performing model is the MLP which consistently provides the best results for all performance measures across most uni and multimodal settings. 
For instance, it provides the best accuracy, F1, and AUROC scores across the audio, A+T, A+V, and A+T+V modalities. 
This is followed by the the Logistic Regression and the SVM Linear models.
Similar to the CMDC dataset, NB performs the worst across most performance measures.
For instance, across the visual modality, NB gave the lowest accuracy, F1, and AUROC scores of $0.53$, $0.53$ and $0.54$ respectively as well as the lowest multimodal (A+T+V) results of $0.58$ across both F1 score and AUROC.
This is followed by KNN. 
%classifier. 
The Decision Tree seems to work better for E-DAIC than CMDC.
%}

\subsubsection{Fairness Evaluation}
Across fairness, several models provide comparable results. 
From Table \ref{tab:RESULTSEDAIC}, we see that the SVM RBF model performs the best for the multimodal A+T+V setup whereas the MLP performs the best for the unimodal audio modality. The Logistic Regression and the Decision Tree models also provide good results across some settings such as the unimodal textual modality.
On the other hand, NB again produces the least fair outcome across most features and modalities. For instance, it produces the least fair outcome across both measures for the unimodal textual, multimodal A+T, T+V, and A+T+V modalities.

%%%%%%%%%%%%%%%%%%%%%%
\section{Discussion of Findings}

%In this section, we discuss our findings and address the RQs as introduced in Section 1.

%%%%%%%%%%%%%%%%%%%%%%
\vspace{2mm}
\noindent\textbf{RQ1: Are there any differences in depression manifestation across gender and countries?}
Results from 
%A possible explanation for why this happened may lie within the MANOVA statistcal analysis undertaken on both datasets (see 
Sections \ref{subsec:CMDCStatisticalAnalysis} indicate that there are significant differences between the different features across depressed vs. non-depressed participants as well as gender for CMDC. 
This is true for both the visual and audio features. 
In addition, the spectral acoustic features and the amplitude acoustic features also varied significantly across gender.
However, this observation is not as significant for E-DAIC 
as highlighted in Section \ref{subsec:EDAICStatisticalAnalysis}.
%The features present in CMDC tended to show more significant differences across depression status than the features present in E-DAIC. 
%which showed almost no significant differences whatsoever. 
%
%In addition, E-DAIC's two-way MANOVA p values (section \ref{subsubsec:TwoWayMANOVAAcousticEDAIC}) are similar to its one-way MANOVA p values (section \ref{subsubsec:MANOVAAcousticEDAIC}), which indicates that there may be very significant differences between the genders in the features. Therefore it is  possible for the classifiers to accidentally learn E-DAIC's well-represented gender characteristics alongside characteristics of depression, potentially explaining why minimal changes materialise in both gender fairness and performance metrics when evaluating unimodal and multimodal setups with E-DAIC.
%
%In sum, 
%to answer \textbf{RQ 1}, 
We see significant differences in depression manifestation across gender in both datasets. 
However, a distinctive feature is that this observation is stronger for CMDC than for E-DAIC. This hints that there may be statistically significant differences in depression manifestation across the sample population in both datasets.

%

%%%%%%%%%%%%%%%%%%%%%%

\noindent\textbf{RQ 2: How do the results compare between different ML models and modalities across the different datasets?}
%
%To answer \textbf{RQ 2}, 
There are differences in ML model performance and fairness. It is interesting to note that for both datasets, certain models such as the SVM Linear and the Logistic Regression models
%seem to 
perform much better, and certain models such as NB perform poorly across both datasets. 
Looking at the gender fairness results on CMDC, 
%it seems that 
multimodal setups generally result in a narrower range of 
%help to reduce the standard deviation of 
$DI$ scores across gender than the unimodal setups.
Using a multimodal setup as 
compared to a unimodal setup 
provides the classifiers with more information to learn from, %compared to a unimodal setup, 
which may have resulted in an improvement across the fairness and performance metrics
%and therefore the fairness and performance metrics improve 
\cite{Mathur_Introducing_2020,Cheong_NotFair_2023}. 
However, %across gender fairness 
for E-DAIC, the range of $DI$ scores remain similar across both unimodal and multimodal setups.

%%%%%%%%%%%%%%%%%%%%%%

\noindent \textbf{RQ3: Are there any differences in ML model performance and fairness across gender and countries? If present, are the differences in performance due to the differences in depression manifestation as hypothesised? }
%
%
%To answer \textbf{RQ 3}, 
We see that overall, the ML models perform better for CMDC than for E-DAIC. In general, model performance was better across all feature-modality combinations when the classifiers are trained and tested on the CMDC dataset. The change in performance between CMDC and E-DAIC is in agreement with Zou et al.'s benchmark evaluations on both datasets \cite{zou_cmdc_2023}. 
It is probable that the classifiers found it more difficult to recognise the defining characteristics of depressed people in E-DAIC given the similarity across E-DAIC's features. 
The contrast %in similarity across features 
between CMDC and E-DAIC could be due to the data elicitation methods employed by the dataset  creators. CMDC's features were derived from 12 questions originating from a semi-structured interview that was clinically approved \cite{zou_cmdc_2023}, whereas E-DAIC's features were derived from casual recorded conversations \cite{ringeval_AVEC_2019, gratch-etal-2014-distress}. 
Perhaps the consistency of subject matter ingrained in CMDC's data made a difference. Alternatively, perhaps depression is outwardly more obvious to spot in China than it is in the USA, and it is  this cultural difference that is factoring into the contrast observed between each dataset's features and its models' performances. This theory would be in agreement with findings from 
%Juhasz et al. 
existing works \cite{Juhasz_cultural_2012,Chang_crossnational_2008,Iwata_race_2002}.
%} and others \cite{
%
%Chang_crossnational_2008,Iwata_race_2002}.

%The MANOVA analysis of the visual and acoustic features in E-DAIC can help illuminate why this pattern is occurring. 

%%%%%%%%%%%%%%%%%%%%%%
\section{Conclusion}
A key takeaway from this study is that there were notable differences between each selected classifier's performance and gender fairness metrics 
%when they were trained and evaluated on 
across the CMDC and E-DAIC datasets. However, it is not conclusive if this is due to cultural differences or due to differences in data collection methods.
%across these datasets. 
%
We could not conduct statistical tests between 
%the data from the two different countries 
both datasets as the data elicitation methods were too different.
%
%Although the difference in data collection methodologies and populations involved 
%
This was one of the big limitations of this study as we could not ascertain if the difference in performance across the two countries were due to a difference in depression manifestation or data collection methodology \cite{zou_cmdc_2023,whang2023data}.
Future work could involve the creation of 
%a consistently sourced dataset across different countries and cultures to facilitate this type of analysis.
%
%This may entail synthesising 
a new multimodal depression dataset where data is collected in multiple countries via a single clinically approved protocol (e.g., semi-structured interview). 
%
\begin{comment}
The data collected from each country could then be combined and a comprehensive evaluation of 
%country-level 
fairness could then be conducted. 
%across different factors.
%just as gender fairness has been evaluated for both datasets in this study.
\end{comment}
%
The limited number of participants %contributing data to both 
in CMDC and E-DAIC was another limiting factor. 
%In the case of CMDC, there were sometimes less than a handful of male participants present in each test partition of each fold in stratified 5-fold cross validation. 
%This is why some of the $DI$ metrics are well-rounded and segmented in Table \ref{tab:RESULTSCMDC}. 
Future work should ensure a reasonable number of samples for the underprivileged group in fairness metric calculation, to experiment with other evaluation methods beyond just stratified k-fold cross validation,
%
%Moreover, 
%in this study of evaluating and comparing ML models' gender fairness metrics, only Equal Accuracy and $DI$ were used as fairness metrics. 
%future work can also 
and consider using other
%There are many more 
fairness metrics such as Equal Opportunity and Equalized Odds \cite{mehrabi_survey_2019} to ensure more thorough and comprehensive evaluation.
%, and future work should look to incorporate these.
Future research can also focus on developing solutions to mitigate bias issues in both datasets. This will enhance our understanding of how to address bias in datasets from different countries. Future work can also investigate other methods \cite{Cheong_2023_WACV, churamani2023towards}
or 
orthogonal forms of fairness \cite{kuzucu2023uncertainty,cheong_neurips}
to better understand the root cause of bias.

%The motivation of our work is to understand how ML performance and fairness differ across datasets and what 
%
%are the challenges that need to be addressed to promote more reliable fairness evaluation.
%
%and development of ML-models for mental wellbeing.
%
We made a conscious effort to repeat the experiments for different algorithms to ensure that the results are not algorithm-dependent.
%
%We identified two main challenges, difference in data curation and potential cultural difference in depression manifestation.
%
\begin{comment}

Our work highlights how 
there is a difference in performance for different depression datasets and there is need 
%for better data curation in order 
channel more efforts to mitigate the risk of bias in the agents or robots powered by these ML algorithms. 
\end{comment}
%
%
We hope that our work will encourage other researchers to take into consideration fairness and other ethical concerns when developing ML-equipped agents and robots for wellbeing \cite{chen2023ai, axelsson2022robots,biondi2023ethical,cheong2024small,cheong2024causal}. 
Given the high-stakes involved in using such agents and robots for wellbeing coaching \cite{churamani2022continual,spitale2023vita,spitale2024appropriateness},
we hope our results and findings will 
encourage more consistent and culturally-aware data collection for depression \cite{lehti2009recognition},
%process in order to address the problem of ML bias in depression detection
%
thus paving the way towards developing fairer 
%depression detection models 
ML-equipped agents and social robots for providing wellbeing
to all.

\section*{Ethical Impact Statement}

We recognise the sensitive nature of this study and have adopted measures aligned with ethical guidelines.
The datasets used have been anonymised by the dataset owners to minimise privacy impact. Our work attempts to avoid any bias against certain groups of people that could result in discrimination. However, our results are limited to the datasets included in this work. 
Future work should repeat the same analysis on other depression datasets to further validate our findings. 

%%%%%%%%%%%%%%%%%%%%%%%%%%%%%%%%%%%%%%%%%%%%%%%%%%%%%%%%%%%%%%%%%%%%%%%%%%%%
\section*{Acknowledgments}
\small
\noindent\textbf{Funding:} J. Cheong is funded by the Alan Turing Institute Doctoral Studentship and the Leverhulme Trust. M. Spitale is supported by PNRR-PE-AI FAIR project funded by the NextGeneration EU program. H. Gunes is supported by the EPSRC/UKRI under grant ref. EP/R030782/1 (ARoEQ). 
\textbf{Open Access:} For open access purposes, the authors have applied a Creative Commons Attribution (CC BY) licence to any Author Accepted Manuscript version arising.
\textbf{Data access:} This study involved secondary analyses of pre-existing datasets. All datasets are described in the text and cited accordingly. 

%\clearpage

\bibliographystyle{IEEEtran}

% Generated by IEEEtran.bst, version: 1.14 (2015/08/26)

%\bibliographystyle{ACM-Reference-Format}
%\bibliography{Jan24}

\end{document}